\documentclass[letterpaper]{article} 
\usepackage{aaai24}  
\usepackage{times}  
\usepackage{helvet}  
\usepackage{courier}  
\usepackage[hyphens]{url}  
\usepackage{graphicx} 
\usepackage{natbib}  
\usepackage{caption} 
\usepackage{algorithm}
\usepackage{algorithmic}
\usepackage{amssymb}
\usepackage{amsmath}

\usepackage{newfloat}
\usepackage{listings}
\DeclareCaptionStyle{ruled}{labelfont=normalfont,labelsep=colon,strut=off} 
\floatstyle{ruled}
\newfloat{listing}{tb}{lst}{}
\floatname{listing}{Listing}
\title{Hyper Association Graph Matching with Uncertainty Quantification for Coronary Artery Semantic Labeling}
\author {
    Chen Zhao\textsuperscript{\rm 1},
    Michele Esposito\textsuperscript{\rm 2},
    Zhihui Xu\textsuperscript{\rm 3},
    Weihua Zhou\textsuperscript{\rm 1}
}
\affiliations {
    \textsuperscript{\rm 1}Department of Applied Computing, Michigan Technological University, Houghton, MI, USA\\
    \textsuperscript{\rm 2}Department of Cardiology, Medical University of South Carolina, Charleston, SC, USA\\
    \textsuperscript{\rm 3}Department of Cardiology, The First Affiliated Hospital of Nanjing Medical University, Nanjing, China\\
    chezhao,whzhou@mtu.edu, espositm@musc.edu,zhihui.xu4@umassmed.edu
}

\usepackage{bibentry}

\begin{document}

\maketitle

\begin{abstract}
Coronary artery disease (CAD) is one of the leading causes of death worldwide. Accurate extraction of individual arterial branches from invasive coronary angiograms (ICA) is critical for CAD diagnosis and detection of stenosis. However, deep learning-based models face challenges in generating semantic segmentation for coronary arteries due to the morphological similarity among different types of coronary arteries. To address this challenge, we propose an innovative approach using the hyper association graph-matching neural network with uncertainty quantification (HAGMN-UQ) for coronary artery semantic labeling on ICAs. The graph-matching procedure maps the arterial branches between two individual graphs, so that the unlabeled arterial segments are classified by the labeled segments, and the coronary artery semantic labeling is achieved. By incorporating the anatomical structural loss and uncertainty, our model achieved an accuracy of 0.9345 for coronary artery semantic labeling with a fast inference speed, leading to an effective and efficient prediction in real-time clinical decision-making scenarios.

\end{abstract}

\section{Introduction}

Coronary artery disease (CAD) has been recognized as a primary cause of mortality worldwide \cite{boden2007optimal}. Invasive coronary angiography (ICA) remains the gold standard for CAD diagnosis \cite{li2015robust}. ICAs are instrumental to aid cardiologists in identifying blockages within the coronary arteries. Nevertheless, it is crucial to acknowledge the limitations inherent in this subjective visual assessment \cite{xian2020main}.

The coronary vascular tree consists of two major systems: the left coronary artery (LCA) and the right coronary artery trees. The LCA is more clinically relevant, as it supplies most of the blood to the left ventricle \cite{parikh2012left}. The LCA system further bifurcates into three main coronary arteries: the left anterior descending (LAD) artery, the left circumflex (LCX) artery, and the left main artery (LMA). The LAD gives rise to diagonal branches (D), while the LCX gives rise to obtuse marginal (OM) branches. Automatically identifying the correct anatomical branches offers valuable insights for the automatic generation of diagnosis reports and quantification of region of interests.

\begin{figure*}[t]
\centering
\includegraphics[width=0.95\textwidth]{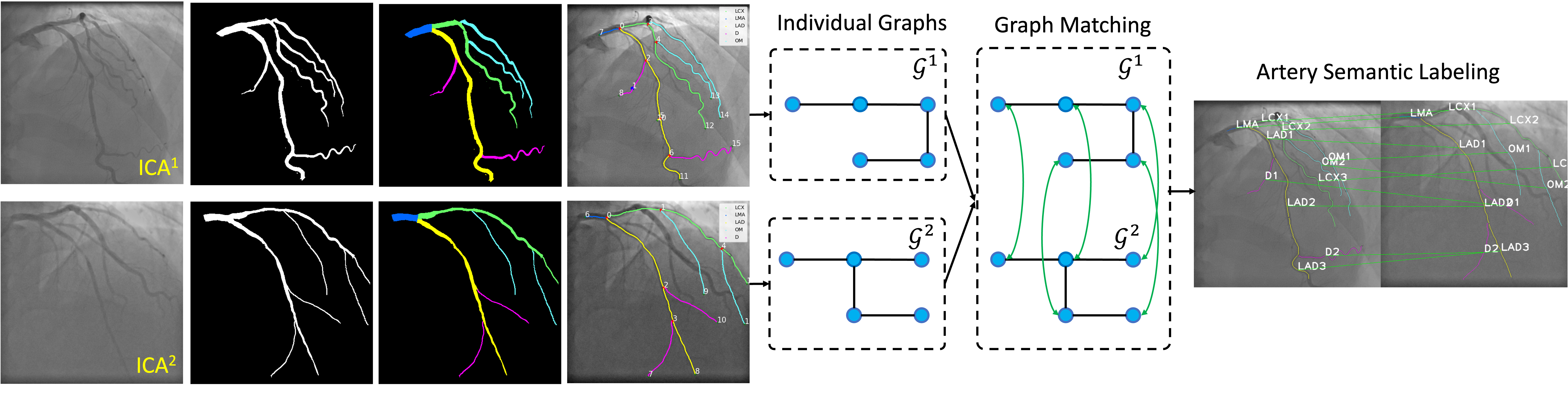} 
\caption{Workflow of HAGMN-UQ for coronary artery semantic labeling. Two ICAs are selected for graph matching. The binary mask is generated by feature pyramid U-Net++ \cite{zhao2021automatic} and the semantic masks are manually annotated and confirmed by experienced interventional cardiologists. The centerlines and key points are extracted and the individual graphs are generated for graph matching.}
\label{fig_workflow}
\end{figure*}

Our proposed approach includes a hyper association graph-based graph-matching network with uncertainty quantification (HAGMN-UQ) to establish semantic correspondences between coronary arterial segments from ICAs. The semantic segmentation problem is transformed into a task of classifying the type of an unlabeled arterial segment by searching for the most similar labeled arterial segment in a template set. The individual graph of the coronary artery tree is generated based on its topology, with each node representing a segment of the coronary artery and each edge indicating the connectivity between arterial segments. A hyper association graph is constructed from two individual graphs, where each vertex corresponds to two nodes in two different individual graphs, representing the correspondence between arterial segments. This transforms the coronary artery semantic labeling task into a vertex classification task using the generated hyper association graph. HAGMN-UQ incorporates an encoder module that embeds vertex and edge features using graph transformer networks, and a decoder module that facilitates feature representation readout. By examining the positive vertices in the association graph, indicating matched nodes from individual graphs, HAGMN-UQ accomplishes the semantic labeling task by assigning mapped labels to coronary arterial segments. The workflow of the proposed HAGMN-UQ for coronary artery semantic labeling is illustrated in Fig. \ref{fig_workflow}.

The highlights and novelty of this paper are shown below:
\begin{itemize}
    \item The utilization of graph matching for coronary artery semantic labeling using hyper association graph.
    \item Node attention and edge attention implemented by the graph transformer network are employed to dynamically aggregate features for graph matching.
    \item Structural loss characterized by the coronary anatomy and the true classification probability loss are employed to perform a trustworthy graph matching, reducing the processing time and improving the accuracy.
\end{itemize}

\section{Related works}

\subsection{Coronary Artery Semantic Segmentation}

Identifying individual coronary arteries in ICAs is challenging due to the morphological similarities among different segments \cite{zhao_agmn}. The projection of the 3D vascular tree into a 2D plane during ICA image acquisition leads to overlaps between arteries, making it difficult to discern intricate structures and boundaries of the coronary arteries \cite{zhang2022progressive}. Various deep learning-based approaches have been proposed for pixel-to-pixel based methods, achieving impressive results in segmenting main arteries \cite{xian2020main, zhang2022progressive, jun2020t}. However, these models focus solely on main branches and may not adequately label side branches, limiting their applicability for comprehensive CAD analysis.

Segment identification-based methods involve vascular tree binary segmentation followed by segment classification. The developed coronary artery semantic labeling using 3D Coronary Computed Tomography Angiography (CCTA) data achieved satisfactory results \cite{cao2017automatic, wu2019automated}; however, they cannot be applied directly to 2D ICAs. Zhao et al. introduced an association graph-based graph-matching network \cite{zhao_agmn} and an edge attention graph matching network \cite{zhao_eagmn} for coronary artery semantic labeling on ICAs. These methods demonstrate the capability to label side branches. However, a limitation lies in the slow inference speed, making it unsuitable for real-time scenarios. This study aims to accurately label coronary arteries and achieve real-time clinical applicability through satisfactory performance and fast inference speed.

\subsection{Graph Matching}

The objective of graph matching is to establish a meaningful correspondence between nodes and edges across different graphs. Graph-matching poses a challenging combinatorial optimization problem based on graph structures \cite{vesselinova2020learning}, which becomes NP-hard or practically infeasible for large-scale scenarios. 

Traditional graph-matching approaches primarily rely on combinatorial techniques, comparing and matching structural components of graphs. In contrast, learning-based methods take a different route by incorporating feature extraction and affinity learning, which excel in handling large-scale and high-dimensional data, and enable the discovery of meaningful patterns and relationships within graphs \cite{khalil2017learning}. Graph neural network (GNN) has further enriched learning-based graph matching. GNN models structured information and transforms the graph-matching problem into a linear assignment task \cite{nowak2018revised, NGM, wang2020combinatorial}.

However, the existing graph-matching algorithms have mainly been explored in domains like nature images, and their application to medical images, with crucial topological features, remains underexplored. Given the significance of topological features in medical image analysis, there is a growing demand for dedicated research to develop graph-matching algorithms tailored specifically to medical imaging. Such algorithms should account for the intricate interactions between structural elements and leverage the inherent knowledge of anatomical connectivity to achieve more accurate and meaningful matching outcomes.

\subsection{Uncertainty Quantification}

Uncertainty quantification (UQ) is a crucial aspect in the optimization of the decision-making process, as it helps mitigate the effects of uncertainties and enhances the reliability of predictions \cite{abdar2021review}. Existing UQ methods have found successful applications in diverse fields, including image classification, image segmentation, and medical applications \cite{baumgartner2019phiseg}. However, despite its widespread application, UQ for graph matching remains understudied. Exploring and developing UQ techniques specific to graph-matching tasks could open new avenues for improving the accuracy and robustness of graph-matching.

\section{Methodology}

This study presents a novel coronary artery semantic labeling method, HAGMN-UQ, focusing on segment identification-based approaches. HAGMN-UQ establishes the node-to-node correspondence between individual graphs while reducing graph-matching inference time. Each node represents a coronary arterial segment, and HAGMN-UQ aims to build one-to-one or one-to-zero mappings for segments in distinct vascular trees, as depicted in Fig. \ref{fig_workflow}.

\subsection{Individual Graph Generation}

FP-U-Net++ \cite{zhao2021automatic} is employed to extract the vascular tree and experienced cardiologists provide pixel-level semantic labels. Arterial centerlines are extracted and key points are identified using an edge-linking algorithm to segment the vascular tree. An individual graph is created, where the nodes represent arterial segments and edges indicate connections. Features from \cite{zhao_agmn}, including topology, pixel, and position features, are used (121 features in total) for artery feature extraction.

\subsection{Hyper Graph Matching for Coronary Artery Semantic Labeling}

Hypergraph is a mathematical structure that generalizes the concept of a graph, where each edge connects multiple nodes simultaneously. Hypergraph matching is an extension of graph matching that allows for matching vertices and hyperedges between individual hypergraphs, exploiting the higher orders between the nodes and edges to provide a robust matching \cite{NGM}. This study concentrates on a 3-uniform hypergraph where each hyperedge contains three nodes and the graph matching network is performed to determine a third-order correspondence between hypergraphs. Formally, the individual 3-uniform hypergraph represented by an attributed undirected graph is defined as $\mathcal{G}=(\mathbb{V}, \mathbb{E}, \mathcal{V}, \mathcal{E})$, where 

\begin{itemize}
    \item $\mathbb{V}=\left\{V_i\right\}$ s.t. $i \in \{1, \cdots, n\}$ represents node set, and $|\mathbb{V}|=n$

    \item $\mathbb{E}=\left\{E_{i j k}\right\}$ s.t. $i,j,k \in \{1, \cdots, n\}$ indicates the set of hyperedges and $|\mathbb{E}|=n_e$.

    \item $\mathcal{V}=\left\{v_i\right\}$ indicates the attribute vectors for each node.

    \item $\mathcal{E}=\left\{e_{i j k}\right\}$ indicates the attribute vectors associated with each hyperedge.
\end{itemize}

\begin{figure*}[t]
\centering
\includegraphics[width=0.9\textwidth]{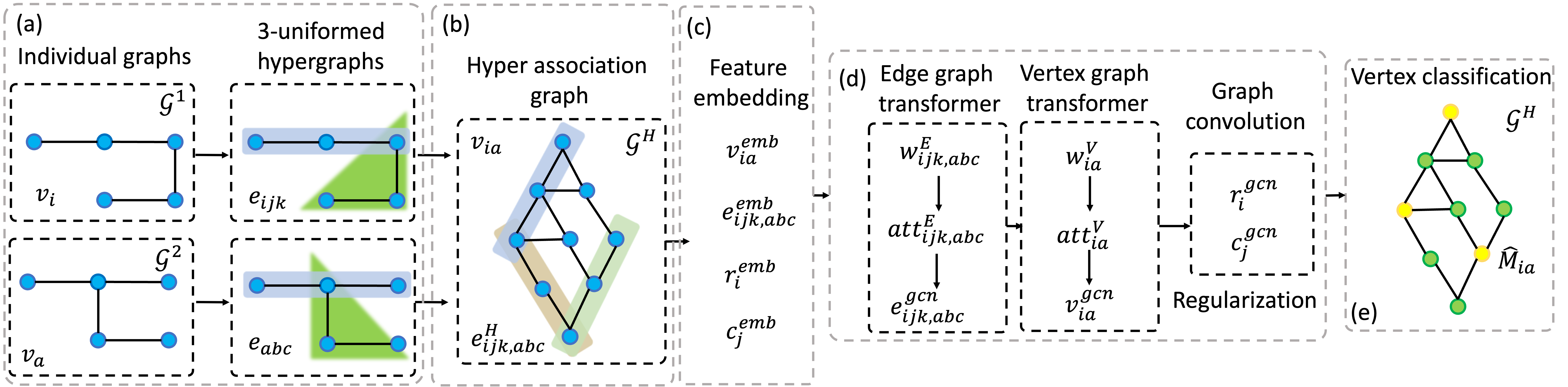} 
\caption{Architecture of HAGMN-UQ. (a) Individual hypergraph generation; (b) Hyper association graph generation; (c) Feature embedding; (d) Feature representation learning; (e) Vertex classification according to the learned feature representations.}
\label{fig_hagmn}
\end{figure*}

A hyperedge is added when physically connected nodes $V_i$, $V_j$, and $V_k$ are adjacent in hypergraphs. Thus, $\mathbb{E}=\{(V_i, V_j, V_k)\} = \{E_{i j k}\}$. Hypergraph matching finds node correspondence between two undirected individual hypergraphs $\mathcal{G}^1=(\mathbb{V}^1, \mathbb{E}^1, \mathcal{V}^1, \mathcal{E}^1)$ and $\mathcal{G}^2=(\mathbb{V}^2, \mathbb{E}^2, \mathcal{V}^2, \mathcal{E}^2)$ using node-to-node and hyperedge-to-hyperedge affinities. Without loss of the generality, it is supposed that $n_1\leq n_2$. The node-to-node correspondence can be represented by an assignment matrix $M\in \{0,1\}^{n_1 \times n_2}$. To eliminate ambiguity, the node in the individual graph is defined as a \textit{node}, while the node in the association graph is defined as a \textit{vertex}. Inspired by the success of applying association graph in combinatorial optimization \cite{NGM}, the hyper association graph is constructed using the individual hypergraphs and the graph matching is transformed into a vertex binary classification task, as shown in Fig. \ref{fig_hagmn}.

\subsubsection{Hyper association graph construction.} Formally, the hyper association graph is denoted as $\mathcal{G}^H=(\mathbb{V}^H, \mathbb{E}^H, \mathcal{V}^H, \mathcal{E}^H)$, where 

\begin{itemize}
    \item $\mathbb{V}^H=\left\{V_i^H\right\}$ s.t. $i \in \{1, \cdots, n_1 \times n_2\}$ represents the node set, and $|\mathbb{V}^H|=n_1 \times n_2$

    \item $\mathbb{E}^H=\left\{E_{i j k, a b c}^H\right\}$ s.t. $i,j,k \in \{1, \cdots, n_1\}$ and $a,b,c \in \{1, \cdots, n_2\}$, indicating the set of edges.

    \item $\mathcal{V}^H=\left\{v_{ia}^H\right\}$  indicates the attributes for each vertex.

    \item $\mathcal{E}^H=\left\{e_{i j k, a b c}^H\right\}$ are the attributes for each hyperedge.
\end{itemize}

Each vertex $V_i^H$ in $\mathcal{G}^H$ combines the two nodes from the two individual hypergraphs. Thus $V^H=\{(V_1^1,V_1^2 ),\cdots,(V_{n_1}^1,V_{n_2}^2)\}$, where $V_i^1$ and $V_a^2$ are nodes from $\mathcal{G}^1$ and $\mathcal{G}^2$, respectively. And each hyperedge in $\mathcal{G}^H$ is derived from two hyperedges in the individual hypergraphs. If the node $V_i^1$, $V_j^1$ and $V_k^1$ from $\mathcal{G}^1$ are connected, and node $V_a^2$, $V_b^2$ and $V_c^2$ from $\mathcal{G}^2$ are connected, then a hyperedge $E_{ijk,abc}=(V_i^1,V_j^1,V_k^1,V_a^2,V_b^2,V_c^2 )$ is constructed in $\mathcal{G}^H$.

The vertex of the hyper association graph is constructed by concatenating the nodes in the individual hypergraphs, and the feature of $v_{ia}^H$ is derived by concatenating the features of $V_i^1$ and $V_a^2$, as shown in Eq. \ref{EQN03}.

\begin{equation}
V_{ia}^H = [v_i, v_a]\in\mathbb{R}^{2d}
    \label{EQN03}
\end{equation}
where $[\cdot]$ indicates concatenation. $d$ indicates the feature dimension. Here, each node is assigned 121 handcrafted features, leading to $d=121$.

The hyperedge features in the individual graph are constructed using the features of the included nodes, as shown in Eq. \ref{EQN04}.

\begin{equation}
e_{i j k}=\left[v_i, v_j, v_k\right] \in \mathbb{R}^{3 d}
    \label{EQN04}
\end{equation}

Similarly, the hyperedge features in the hyper association graph are constructed by concatenating the features of hyperedges in the individual hypergraphs, as shown in Eq. \ref{EQN05}.

\begin{equation}
e_{i j k, a b c}^H=\left[e_{i j k}, e_{a b c}\right] \in \mathbb{R}^{6 d}
    \label{EQN05}
\end{equation}

\subsubsection{$L^1$-Norm constraint regularization.} The core spectral hypergraph matching issue is its uniform solution \cite{duchenne2011tensor}, implying challenges in obtaining a precise 0 and 1 assignment matrix in $M$. As in Liao et al.'s hypergraph matching \cite{liao2021hypergraph}, we adopt row and column hyperedge constraints. Formally, the hyper association graph is converted into $\mathcal{G}^H=(\mathbb{V}^H, \mathbb{E}^H, \mathbb{R}^H, \mathbb{C}^H, \mathcal{V}^H, \mathcal{E}^H, \mathcal{R}^H, \mathcal{C}^H)$, where $\mathbb{R}^H$ and $\mathbb{C}^H$ indicate the row and column hyperedge; $\mathcal{R}^H \in \mathbb{R}^1$ and $\mathcal{C}^H \in \mathbb{R}^1$ represent features of row and column hyperedges; and they are initialized as zeros. In addition, $\mathcal{R}^H=\{r_i^H\}$ s.t.  $i\in \{1,\cdots, n_1\}$ and $\mathcal{C}^H=\{c_j^H\}$ s.t. $j\in \{1,\cdots,n_2\}$.

\subsubsection{Feature embedding.} The proposed HAGMN-UQ first performs feature representation learning using multilayer perceptron (MLP). Formally, four MLPs are employed to perform feature embedding, as defined in Eq. \ref{EQN06} to Eq. \ref{EQN09}.

\begin{equation}
v_{i a}^{e m b}=f_v^{e m b}\left(v_{i a}^H\right) 
    \label{EQN06}
\end{equation}

\begin{equation}
e_{i j k, a b c}^{e m b}=f_e^{e m b}\left(e_{i j k, a b c}^H\right) 
    \label{EQN07}
\end{equation}

\begin{equation}
r_i^{e m b}=f_r^{e m b}\left(r_i^H\right)
    \label{EQN08}
\end{equation}

\begin{equation}
c_j^{e m b}=f_c^{e m b}\left(c_j^H\right)
    \label{EQN09}
\end{equation}
where $f_v^{emb}$, $f_e^{emb}$, $f_r^{emb}$ and $f_c^{emb}$ are MLPs for vertex, hyperedges, row and column regularization embeddings, respectively.

\subsubsection{Hyperedge graph convolution using graph transformer.} After performing feature representation learning, the graph transformer \cite{dwivedi2020generalization} is employed to update the edge feature first. The core idea of graph convolution networks (GCN) lies in performing graph convolutions by equally pooling information from neighboring nodes and edges. However, for topology-based medical image semantic labeling, distinct artery types contribute uniquely to the central artery segment. Employing the self-attention mechanism from the Transformer \cite{vaswani2017attention}, the graph transformer processes nodes and edges, capturing inter-element dependencies and enabling dynamic neighbor integration for graph-level computations. 

The edge graph transformer (EGT) is employed to aggregate features, where the hyperedge-embedded features serve as queries, while vertex-embedded features act as keys and values for attention. The calculated attention weights and the vertex attention values for EGT are shown in Eqs. \ref{EQN10} and \ref{EQN11}.

\begin{equation}
w_{i j k, a b c}^E = \sigma \left(a_e^Q\left(e_{i j k, a b c}^{e m b}\right) \cdot \sum_{\forall V_{m l}} \frac{a_e^K\left(v_{m l}^{e m b}\right)}{\left|N_v(E_{i j k, a b c})\right|}\right) 
    \label{EQN10}
\end{equation}

\begin{equation}
a t t_{i j k, a b c}^E=w_{i j k, a b c}^E \cdot \sum_{\forall V_{m l}} \frac{a_e^V\left(v_{m l}^{e m b}\right)}{\left|N_v\left(E_{i j k, a b c}\right)\right|}
    \label{EQN11}
\end{equation}
where $\sigma$ denotes softmax classification; $a_e^Q$, $a_e^K$ and $a_e^V$ are MLPs for multi-head attention; $a_e^Q(e_{ijk,abc}^{emb})$, $a_e^K(v_{ia}^{emb})$ and $a_e^V(v_{ia}^{emb})$ are the queries, keys and values for edge feature attention, respectively. $N_v\left(E_{i j k, a b c}\right)$ represents the set of the connected vertices for hyperedge $E_{i j k, a b c}$ in $\mathcal{G}^H$ and $|N_v\left(E_{i j k, a b c}\right)|$ is the number of vertices in the set. In other words, $V_{ml} \in N_v\left(E_{i j k, a b c}\right)$. Then, the features of hyperedges are updated in Eq. \ref{EQN12}.

\begin{equation}
e_{i j k, a b c}^{g c n}=g_e^{g c n}\left(\left[e_{i j k, a b c}^{e m b}, a t t_{i j k, a b c}^E\right]\right)
    \label{EQN12}
\end{equation}
where $g_e^{gcn}$ is an MLP. 

Then, the vertex features are updated using a vertex graph transformer (VGT), which dynamically integrates the features of the connected hyperedges of the vertex. The embedded features of vertices serve as queries, and the features of hyperedges act as keys and values. Formally, the calculated attention weights and edge attention values for VGT are shown in Eqs. \ref{EQN13} and \ref{EQN14}.

\begin{equation}
w_{i a}^V=\sigma \left(a_v^Q\left(v_{i a}^{e m b}\right) \cdot \sum_{\forall E_{n m l, n^\prime m^\prime l^\prime}} \frac{a_v^K\left(e_{n m l, n^\prime m^\prime l^\prime}^{g c n}\right)}{\left|N_e\left(V_{i a}\right)\right|}\right)
    \label{EQN13}
\end{equation}

\begin{equation}
    a t t_{i a}^V=w_{i a}^V \cdot \sum_{\forall E_{n m l, n^\prime m^\prime l^\prime}} \frac{a_v^V\left(e_{n m l, n^\prime m^\prime l^\prime}^{g c n}\right)}{\left|N_e\left(V_{i a}\right)\right|}
    \label{EQN14}
\end{equation}
where $a_v^Q$, $a_v^K$ and $a_v^V$ are MLPs for multi-head attention; $N_e(V_{ia})$ represents the set of connected hyperedges for vertex $V_{ia}$ in $\mathcal{G}^H$ and $|N_e(V_{ia})|$ is the number of hyperedges in the set. Similarly, $E_{n m l, n^\prime m^\prime l^\prime} \in N_e\left(V_{i a}\right)$. Then, the features for vertices are updated according to Eq. \ref{EQN15}.

\begin{equation}
v_{i a}^{g c n}=g_v^{g c n}\left(\left[v_{i a}^{e m b}, a t t_{i a}^V\right]\right)
    \label{EQN15}
\end{equation}

For the row and column regularization terms, the plain GCN is employed to update the features in Eqs. \ref{EQN16} and \ref{EQN17}.

\begin{equation}
r_i^{g c n}=g_r^{e m b}\left(\left[r_i^{e m b}, \sum_j^{n_2} \frac{v_{i j}^{g c n}}{n_2}\right]\right)
    \label{EQN16}
\end{equation}

\begin{equation}
c_j^{g c n}=g_c^{e m b}\left(\left[c_j^{e m b}, \sum_i^{n_1} \frac{v_{i j}^{g c n}}{n_1}\right]\right)
    \label{EQN17}
\end{equation}

\subsubsection{Iterative Graph Convolution.} Graph convolution iterates on the embedded hyper association graph for $N_{mp}$ rounds. This repetition enhances information flow and captures higher-order dependencies. Such iteration fosters a deeper and more nuanced comprehension of graph-structured data,  leading to improved performance in graph matching.

\subsubsection{Decoder and Classification.} A decoder layer is employed to transform the embedded vertex features in Eq. \ref{EQN15} into classification probabilities with a Softmax classifier. Formally, the decoder layer is represented by an MLP and the predicted vertex class is shown in Eq. \ref{EQN18}.

\begin{equation}
\hat{y}_{ia} = \sigma (f_v^{dec}(v_{ia}^{gcn}))
    \label{EQN18}
\end{equation}
where $\hat{y}_{ia} \in \mathbb{R}^2$ indicates the one-hot encoded probability that the vertex $V_i$ in $\mathcal{G}^1$ matches the vertex $V_a$ in $\mathcal{G}^2$. $f_v^{dec}$ is the decoder implemented by MLPs.

\subsubsection{Optimization.} The node-to-node correspondence between the vertex classification results and the ground truth is used to optimize the model. The ground truth and classification results are denoted as two permutation matrices $M \in \mathbb{R}^{n_1 \times n_2}$ and $\hat{M} \in \mathbb{R}^{n_1 \times n_2}$, where the element $M_{ia}$ indicates the relationship between node $V_i^1\in G^1$ and node $V_a^2 \in G^2$, and $\hat{M}_{ia}=\text{argmax} \hat{y}_{ia}$. The permutation loss \cite{NGM}, computed by cross-entropy between predicted vertex class and ground truth, serves as the objective function, as shown in Eq. \ref{EQN19}.

\begin{equation} \label{EQN19}
\mathcal{L}_{perm} =-\sum_{i=1}^{n_1} \sum_{a=1}^{n_2}\left(1-\hat{M}_{i a}\right) \log \left(1-M_{i a}\right) +\hat{M}_{i a} \log M_{i a}
\end{equation}

To explicitly reflect the confidence of the graph-matching and further accelerate the inference procedure, the true class probability (TCP) \cite{corbiere2019addressing} is employed to quantify the classification confidence. TCP uses the softmax output probability corresponding to the real label as confidence. Formally, TCP for vertex $V_{ia}^H$ is defined in Eq. \ref{EQN20}.

\begin{equation} \label{EQN20}
TCP^{ia} = M_{ia} \cdot \hat{y}_{ia}^{M_{ia}}
\end{equation}
where $\hat{y}_{ia}^{M_{ia}}$ indicates $M_{ia}$-th element of $\hat{y}_{ia}$. A confidence neural network implemented using MLP is introduced to approximate $TCP^{ia}$, which is denoted as $f_v^c$. The approximated TCP is defined in Eq. \ref{EQN21}.

\begin{equation} \label{EQN21}
\widehat{TCP}^{ia} = f_v^c(f_v^{dec}(v_{ia}^{gcn}))
\end{equation}

The confidence loss is defined as the mean squared error between the TCP and the approximated TCP in Eq. \ref{EQN22}. And the overall loss function is defined in Eq. \ref{EQN23}.

\begin{equation} \label{EQN22}
\mathcal{L}_{TCP}=\sum_{i=1}^{n_1} \sum_{a=1}^{n_2} (\widehat{TCP}^{ia}-TCP^{ia})^2
\end{equation}

\begin{equation} \label{EQN23}
\mathcal{L}=\mathcal{L}_{perm} + \alpha \mathcal{L}_{TCP}
\end{equation}
where $\alpha$ is a hyperparameter balancing the weights between the graph matching loss and the confidence loss.

\subsection{Training and Testing}

The dataset is split into three subsets, including $D_{tr}$ for training, $D_{te}$ for testing, and $D_{tp}$ for the template set. Experienced cardiologists labeled each arterial segment. By leveraging the labeled data, the node correspondences between arterial segments are automatically identified, and a ground truth permutation matrix $M$ is generated. The pairs of individual graphs from $D_{tr}$ are chosen, and a hyper association graph is generated based on these selected pairs to train the HAGMN-UQ using Eq. \ref{EQN23}. It is noted that ICAs are chosen from the same view angle due to variations in the anatomy and visual characteristics of the coronary arteries between these specific view angles.

During the testing, $\mathcal{G}^1$ represents the tested ICA from $D_{te}$, and $\mathcal{G}^2$ is the template ICA from $D_{tp}$. The hyper association graph is constructed and the permutation matrix $\hat{M}$ is calculated. The Hungarian algorithm \cite{kuhn1955hungarian} is further applied to discretize $\hat{M}$ and an assignment matrix is generated. 

The previously developed graph matching-based methods for coronary artery semantic labeling \cite{zhao_agmn, zhao_eagmn} are limited by the relatively slow prediction during the testing, as each tested ICA in $D_{te}$ is compared with every ICA in $D_{tp}$. In this paper, a structural loss characterized by the coronary artery anatomy and the approximated TCP in Eq. \ref{EQN21} are employed to perform UQ so that reduces the number of the compared graphs and accelerates the inference speed. 

\textbf{Structural Loss}. In the coronary artery system, the coronary arteries follow strict topology based on cardiovascular anatomy. For example, the first parts of LCX and LAD are connected to LMA, the side D branches are connected to LAD and the side OM branches are connected to LCX. If the predicted arterial labels don't follow the anatomy, a weighted penalty is applied to the graph-matching results. We designed a look-up table that stores the physical connectivity between the arterial branches. If the connected arteries of the predicted artery are identical to the arterial branches in the look-up table, then the prediction is confident. In contrast, if not all the connected branches of the predicted artery are identical to the arteries in the look-up table, we calculated the percentage of the error connections as the weight to penalize the prediction. Formally, the penalty weight is denoted as $T_e^{i}$ for $i$-th node in the tested ICA.

\textbf{Test using TCP}. The error weight in structural loss is multiplied by approximated TCP in Eq. \ref{EQN21} as the adjusted confidence, denoted as $T_e^i\cdot \widehat{TCP}^i$. If the adjusted confidence is greater than the confidence threshold, $T_c$, then we accept this prediction. If each node in the tested ICA has a confident match, the testing procedure is terminated. The remaining ICAs in the $D_{tp}$ are no longer needed, reducing the number of the compared ICAs and accelerating the graph matching during the inference. 

\begin{algorithm}[tb]
\caption{Testing process of our proposed HAGMN-UQ.}
\label{alg:algorithm1}
\textbf{Input}:  \\
$\mathcal{G}_{te}$: Tested individual graph with $n_1$ nodes\\
$D_{tp}=\{\mathcal{G}_{tp}^1,\cdots \mathcal{G}_{tp}^{N_{tp}}\}$: Template set \\
$T_c$: confidence threshold \\
\textbf{Output}: Labels for nodes in $\mathcal{G}_1$.
\begin{algorithmic}[1] 
\STATE $A=\{False, \cdots, False\}$ with $n_1$ elements.
\FOR{$i \in \{1, \cdots, N_{tp}\}$}
\IF {$\mathcal{G}_{te}$ and $\mathcal{G}_{tp^i}$ are from the same view angle}  
\STATE Construct hyper association graph $\mathcal{G}^H$.
\STATE Calculate assignment matrix $\hat{M}$ by HAGMN-UQ.
\STATE Discretize $\hat{M}$ by Hungarian algorithm.

\FOR{$j \in \{1, \cdots, n_1\}$}
    \IF{No structural loss presented in $V_j$}
        \STATE $A_j=True$
    \ENDIF
\ENDFOR

    \IF {$\forall a \in A$ is True}
        \STATE \textbf{break}. \# Terminate graph matching
    \ELSE 
        \FOR{$j \in \{1, \cdots, n_1\}$}
            \IF {$T_e^j\cdot \widehat{TCP}^j \geq T_c$ \textbf{and} $A_j=False$} 
                \STATE $A_j=True$
            \ENDIF
        \ENDFOR
    \ENDIF
\ENDIF
\ENDFOR
\STATE \textbf{return} Labels for nodes in $\mathcal{G}_1$.
\end{algorithmic}
\end{algorithm}

The designed algorithm for testing using our HAGMN-UQ with confidence loss is shown in Algorithm \ref{alg:algorithm1}. The template graphs in $D_{tp}$ are sequentially selected for graph matching with $\mathcal{G}_{te}$. Each element in the boolean vector $A$ defined in Line 1 indicates whether the node in $\mathcal{G}_{te}$ has been matched with high confidence. During the testing, if the node $V_j \in \mathcal{G}_{te}$ has no structural loss, then we accept the prediction, as shown in Lines 7. For the node with structural loss, the approximated TCP multiplied by the weights of the structure loss is used to determine if the prediction is confident, as shown in Lines 15 to 19. If all nodes have been assigned a confidence label, then the testing is terminated. Otherwise, the next graph in $D_{tp}$ is selected for graph matching with $\mathcal{G}_{te}$, as the for loop in Line 2. 

\subsection{Evaluation}

The coronary artery semantic labeling problem is converted into a multi-class classification problem among arterial segments. The accuracy (ACC), macro precision (PREC), macro recall (REC), and macro F1-score (F1) are used to evaluate the model performance.

\section{Experiments}

\subsection{Dataset} In this study, 263 and 455 ICAs from Site 1 and Site 2 are enrolled, respectively. All the images were resized into $512 \times 512$. Table \ref{TABLE01} shows the number of images in each view angle. 

\subsection{Implementation details} 
We implemented the proposed HAGMN-UQ using TensorFlow 2 and the grid search was performed to fine-tune the optimal hyperparameters. The MLPs in Eq. \ref{EQN06} to Eq. \ref{EQN18} were set as 2 with 64 neurons with a ReLU activation function. The number of graph convolution, $N_{mp}$, was set as 2. The $\alpha$ in Eq. \ref{EQN23} was set as 0.1. The UQ threshold, $T_c$, was set as 0.4. An Adam optimizer with a learning rate of $0.0001$ was employed to train the HAGMN-UQ.

\begin{table}[t]
\centering
\begin{tabular}{l|l|llll}
    \hline
    Site & View angle & LAO & RAO & AP & TOTAL \\  \hline
    Site 1 & CRA & 42 & 19 & 18 & 79 \\
    Site 1 & CAU & 18 & 116 & 56 & 190 \\ \hline
    Site 2 & CRA & 44 & 16 & 123 & 183 \\
    Site 2 & CAU & 20 & 220 & 26 & 266 \\ \hline
    Total & CAU & 124 & 371 & 223 & 718 \\
     \hline
\end{tabular}
\caption{View angles and number of enrolled subjects}
\label{TABLE01}
\end{table}

\subsection{Coronary Artery Semantic Labeling Performance}

10\% of the ICAs were randomly selected as $D_{tp}$ and the rest ICAs were used for 5-fold cross-validation. The performance on the five-folds was reported in Table \ref{TABLE02}. As illustrated in Table \ref{TABLE02}, the proposed HAGMN-UQ achieved an ACC Of 0.9345, indicating 93.45\% of the arterial segments were correctly classified during the testing. 

\begin{table}[t]
\centering
\begin{tabular}{l|llll}
    \hline
        Artery Type & ACC & PREC & REC & F1 \\ \hline
        LMA & .9960 & .9979 & .9960 & .9969 \\ 
        LAD & .9550 & .9517 & .9556 & .9536 \\ 
        LCX & .9210 & .9250 & .9217 & .9234 \\ 
        D   & .9350 & .9313 & .9361 & .9337 \\ 
        OM  & .8830 & .8890 & .8841 & .8865 \\ 
        weighted avg & .9345 & .9390 & .9387 & .9389 \\ \hline
    \end{tabular}
\caption{Achieved performance for coronary artery semantic labeling using HAGMN-UQ}
\label{TABLE02}
\end{table}

We also compared the performance of the proposed HAGMN-UQ with the existing segment-based coronary artery semantic labeling and other graph-matching methods, including

\begin{itemize}
    \item BiTL \cite{cao2017automatic}: The bidirectional tree LSTM (BiTL) is initially applied for coronary artery semantic labeling using CCTA. We adopted the same network architecture but extracted the spatial locations and directions of coronary
    arteries in 2D.

    \item IPCA \cite{wang2020combinatorial}. Iterative Permutation Cross-graph Affinity (IPCA) is designed for graph matching by comparing node features between individual graphs iteratively with cross-graph feature embedding.

    \item NGM \cite{NGM}. The neural graph-matching network (NGM) employs the association graph-induced affinity matrix for graph matching.

    \item AGMN \cite{zhao_agmn}: association graph-based graph matching network is developed for coronary artery semantic labeling based on association graph using ICAs.

    \item EAGMN \cite{zhao_eagmn}: edge attention graph-based graph matching network is an extension of AGMN by adding the edge attention mechanism.

\end{itemize}

For each baseline, we performed the 5-fold cross-validation using the enrolled 718 ICAs. For the graph matching-based networks, we also split the enrolled subjects into a $D_{tr}$, $D_{te}$ and $D_{tp}$. However, during the testing, each tested ICA was compared with every ICA in $D_{tp}$. The performance comparisons are shown in Table \ref{TABLE_COMP}.

\begin{table}[t]
\centering

\begin{tabular}{l|llllllll}
\hline
Model      & LMA  & LAD  & LCX  & D    & OM   & ACC \\ \hline
BiTL       & .600 & .938 & .677 & .739 & .532 & .7291 \\ 
AGMN       & .990 & .873 & .864 & .832 & .808 & .8639 \\ 
EAGMN      & .994 & .893 & .884 & .851 & .796 & .8767 \\
IPCA       & \textbf{.998} & .911 & .894 & .896 & .839 & .9003 \\ 
NGM        & .988 & .922 & .896 & .900 & .840 & .9039 \\
ours       & .996 & \textbf{.955} & \textbf{.921} & \textbf{.935} & \textbf{.883} & \textbf{.9345} \\ \hline
        
\end{tabular}
\caption{Comparison of coronary artery semantic labeling between different models in ACC.}
\label{TABLE_COMP}
\end{table}

Table \ref{TABLE_COMP} displays that HAGMN-UQ achieved the highest accuracy of 0.9345 for coronary artery semantic labeling. BiTL, originally designed for 3D CCTA datasets, is unsuitable for ICAs, underlining the challenge of labeling coronary arteries from 2D images. For the graph matching-based methods, IPCA first performs feature embedding using individual graphs, and then performs cross-graph feature embeddings using GCNs. Compared to association graph-based graph matching algorithms, individual graph-based matching methods are computationally efficient and suitable for simpler graphs but may struggle with complex graph structures. NGM first embeds features of nodes of the individual graphs to calculate the affinity matrix and uses the calculated affinity matrix as the adjacency matrix for the association graph. Then, NGM performs the feature embedding according to the node features for vertex classification.

On the contrary, AGMN, EAGMN and HAGMN-UQ concatenate the node features of the individual graphs and perform the vertex embedding using the concatenated features. The adjacency matrix of the association graph is generated by the connectivity of the individual graphs rather than the affinity matrix used by NGM. HAGMN-UQ also incorporates high-order edge affinity using hyperedges. Experimental results indicate that the anatomical connections between different arterial segments and high-order affinity provides valuable cues and constraints that guide the matching process. It helps ensure that the matched arterial segments are not only similar in appearance or position but also consistent with the underlying anatomical structure. HAGMN-UQ achieved the highest performance with an ACC of 0.9345 for all types of artery labeling, indicating the superiority of the usage of hypergraph and uncertainty quantification for graph matching.

\subsection{Computation Complexity}

For model complexity evaluation, we compared inference time, weight count, averaged compared graphs count during inference, and total required Floating-Point Operations (FLOPs) in Table \ref{TABLE_COMPLEX}. All tests were performed on an RTX 3090 workstation.

\begin{table}[t]
\centering

\begin{tabular}{l|lllll}
\hline
Model  & Time & \# Weights  & \# GM  & FLOPs \\ \hline
BiTL   & .016       &  25.793K    & 1.000  &  129.0K \\ 
AGMN   & .551      & 296.612K    & 17.744 & 16.9M  \\ 
EAGMN  & .659      & 301.380K    & 17.744 & 13.5M  \\ 
IPCA   & .202      &  40.512K    & 17.744 & 904.9M \\ 
NGM    & .295      &  66.181K    & 17.744 & 2501.9M\\ 
ours   & .102      & 234.033K    & 1.085  & 432.9M\\ \hline
        
\end{tabular}
\caption{Comparison of the computation complexity. The inference time (in seconds), number of weights, averaged number of compared graphs for each tested graph, and the required total FLOPs are presented.}
\label{TABLE_COMPLEX}
\end{table}

In HAGMN-UQ, the averaged compared individual graphs in $D_{tp}$ was 1.085 during testing until filling the assignment vector in Line 1 of Algorithm \ref{alg:algorithm1}, substantially reducing prediction time. Conversely, in other graph-matching-based algorithms (e.g., AGMN, EAGMN, NGM, IPCA), all individual graphs in $D_{tp}$ are utilized for graph matching, and final predictions are obtained through majority voting. As a result, the computation complexity of the proposed HAGMN-UQ was significantly reduced, signifying the enhancement of decision-making confidence through UQ incorporation in graph matching. This, in turn, results in efficient predictions in clinical decision-making scenarios, rendering graph matching for coronary artery semantic labeling more feasible for real-time clinical applications. IPCA, NGM, and HAGMN-UQ all outperformed BiTL, AGMN, and EAGMN, with HAGMN-UQ requiring 432.9M FLOPs, which was lower than the demands of IPCA and NGM. The average inference time per image stood at 0.102 seconds, affirming its practicability in clinical settings.

\subsection{Ablation Study}

The proposed HAGMN-UQ for coronary artery semantic labeling includes three major modules, EGT, VGT and UQ. To verify the effectiveness of the designed modules and the proposed testing algorithm, we built the models with and without using EGT and VGT, and performed the graph matching with and without using UQ. The results of the ablation study are shown in Table \ref{TABLE_ABA}. 

\begin{table}[t]
\centering
\begin{tabular}{lll|llll}
    \hline
VGT & EGT & UQ & ACC & PREC & REC & F1 \\ \hline
 &  &                      & .9070 & .9123 & .9123 & .9121 \\
 &  & \checkmark           & .9245 & .9292 & .9295 & .9293 \\
 & \checkmark &            & .9068 & .9110 & .9103 & .9104 \\
 & \checkmark & \checkmark & .9274 & .9321 & .9322 & .9321 \\
\checkmark &  &            & .8991 & .9046 & .9041 & .9040 \\
\checkmark &  & \checkmark & .9253 & .9300 & .9300 & .9300 \\
\checkmark & \checkmark &  & .9123 & .9164 & .9159 & .9158 \\
\checkmark & \checkmark & \checkmark & \textbf{.9345} & \textbf{.9390} & \textbf{.9387} & \textbf{.9389} \\ \hline
    \end{tabular}
\caption{Ablation results of the employment of EGT, VGT and UQ.}
\label{TABLE_ABA}
\end{table}

According to Table \ref{TABLE_ABA}, integrating VGT and EGT simultaneously improved the model's performance in ACC. Additionally, incorporating the UQ strategy proposed in Algorithm \ref{alg:algorithm1} significantly enhanced the ACC by 2.22 \% (from 0.9123 to 0.9345). For each baseline model, using the UQ strategy improved the performance by approximately 2\% in ACC. These results indicate that VGT and EGT work effectively together, and the incorporation of UQ is crucial in making final decisions.

\section{Conclusion}

In this paper, we present the development and validation of a novel algorithm for coronary artery semantic labeling in ICAs, which exhibits both high accuracy and fast inference speed. The results obtained from our research showcase the promise of our HAGMN-UQ algorithm in computer-aided image analysis for interventional cardiology.

\bibliography{aaai24}

\end{document}